\documentclass[fleqn,10pt]{wlscirep}
\usepackage[utf8]{inputenc}
\usepackage[T1]{fontenc}
\usepackage[numbers]{natbib}
\usepackage{tabularx}
\usepackage{multirow}
\usepackage{pdflscape}
\usepackage{afterpage}
\usepackage{makecell}
\newcolumntype{C}[1]{>{\centering\arraybackslash}p{#1}}
\newcolumntype{L}[1]{>{\raggedright\arraybackslash}p{#1}}

\title{Data Leakage Inflates Generalizability of Power Outage Prediction Models}

\author[1,*]{Yamil Essus}
\author[2]{Ranga Raju Vatsavai}
\author[3,4]{Benjamin Rachunok}
\affil[*]{yamil.essus@utoronto.ca}
\affil[1]{Department of Civil and Mineral Engineering, University of Toronto, Toronto, Canada}
\affil[2]{Computer Science Department, North Carolina State University, Raleigh, NC}
\affil[3]{Department of Industrial and Systems Engineering, North Carolina State University, Raleigh, NC}
\affil[4]{Operations Research Program, North Carolina State University, Raleigh, NC}

\begin{abstract}
Power outage prediction models are increasingly used in assessments of climate-driven infrastructure risk, yet current evaluation practices obscure whether these models generalize to the novel conditions such applications require. We identify three common methodological choices in power outage prediction models that influence their ability to generalize across spatial, temporal, and event-based settings. We compare the predictive performance impacts of different methodological decisions using publicly available data for the U.S. East Coast from 2018 to 2023 and feature sets derived from weather reanalysis and land-cover data, and embeddings from a GeoAI foundation model (Prithvi WxC). Specifically, we assess model performance under multiple test selection strategies, including unfiltered random splits, leave-one-state-out, and leave-one-event-out designs, which increasingly approximate real-world deployment conditions. While random train-test splits yield strong performance, we show that these results are inflated by spatial and temporal autocorrelation. Under spatial and temporal holdout experiments, predictive accuracy degrades substantially, with models often failing to outperform a simple null baseline. Incorporating GeoAI foundation model embeddings yields limited and inconsistent improvements, primarily for spatial generalization, and does not resolve poor event-level transferability. These findings suggest that, given current data availability and evaluation practices, publicly trained outage prediction models offer limited and uncertain operational value. Progress will likely require improved data coverage, more realistic evaluation protocols, and a shift in focus from marginal modeling advances toward addressing structural data constraints.
\end{abstract}
\begin{document}

\flushbottom
\maketitle
%
%
\thispagestyle{empty}



\section*{Introduction}
Power outages regularly cause alarming quality-of-life and economic losses. Accordingly, predicting outages based on environmental, geographic, and infrastructure data using machine learning techniques has been an active area of research for the last 15 years
\citep{nateghiComparisonValidationStatistical2011,
guikemaPredictingHurricanePower2014,
mcrobertsImprovingHurricanePower2018, shashaaniMultiStagePredictionZeroInflated2018, cerraiPredictingStormOutages2019, cerraiOutagePredictionModels2020, tervoShortTermPredictionElectricity2019,tervoPredictingPowerOutages2021,watsonWeatherrelatedPowerOutage2020, nateghiMultiDimensionalInfrastructureResilience2018, yangQuantifyingUncertaintyMachine2020, yangEffectLeadTimeWeather2021, aroraProbabilisticMachineLearning2023,
udehProbabilisticStormElectric2024}. Broadly, most existing studies aim to find spatially and temporally generalizable relationships between covariates and power outages by training statistical learning or machine learning models. Generalizable, predictive approaches are necessary due to climate change, which has increased the severity and frequency of adverse weather events outside of historical records \citep{xuResilienceRenewablePower2024,xieRoleElectricGrid2024,tarrojaQuantifyingClimateChange2016,ridhaClimateChangeImpacts2022,neumannClimateChangeRisks2015,guddantiComprehensiveReviewImpacts2025}, meaning previous patterns of outages are likely not reflective of future conditions.

The role of power outage prediction models is evolving in response to both methodological and environmental changes. On the methodological side, advances in geospatial artificial intelligence (GeoAI) and foundation models have enabled the use of large-scale, high-dimensional datasets to train models across diverse regions. On the environmental side, climate change is increasing the prevalence of extreme events outside the range of historical observations, reducing the reliability of models that rely on stationary relationships. Together, these developments place a new emphasis on model generalizability. This necessitates an increased focus on applying more stringent evaluation protocols to assess the robustness of model performance across diverse spatial, temporal, and event conditions. 

In this context, we identify three common methodological choices in power outage prediction models that influence their ability to generalize across spatial, temporal, and event-based settings. First, existing work predominantly relies on the use of random split test selection, which potentially introduces data leakage due to the spatio-temporal correlations in the weather covariates and leads to inflated performance metrics \citep{Roberts_Bahn_Ciutietal_2017, plotonSpatialValidationReveals2020, kapoorLeakageReproducibilityCrisis2023}. Second, reliance on absolute impact metrics (e.g., number of customers or households without power) as the target variable, which conflates storm severity with regional population size and thereby constrains generalizability to new areas. Lastly, testing data in prior studies is almost exclusively drawn from the same geographic regions used for training, and the outage events selected for evaluation are typically manually chosen, limiting the diversity of conditions under which these models are tested and obscuring whether reported performance metrics reflect true predictive capacity. While these issues do not invalidate the findings of previous studies, they do limit their application to the narrow spatial and temporal extent in which they were trained. Other disciplines, such as ecology \citep{Roberts_Bahn_Ciutietal_2017, valaviBlockCVPackageGenerating2019, plotonSpatialValidationReveals2020}, satellite imagery \citep{nalepaValidatingHyperspectralImage2019}, and medicine \citep{boneApplyingMachineLearning2015, poldrackEstablishmentBestPractices2020, vandewieleOverlyOptimisticPrediction2021}, have recently undertaken similar efforts to refine evaluation practices to better assess model generalizability.

Failing to recognize poor model generalization has the potential to increase outage risk. Spatially and temporally inaccurate estimations of outages result in delayed responses and longer outages, which lead to cascading effects over many interdependent systems, including water supply, communication networks, and transportation \citep{Castillo_2014, essusElectricVehiclesLimit2024}, and heavy disruptions on essential systems, such as healthcare and agriculture \citep{Ulak_Kocatepe_Konila_Sriram_Ozguven_Arghandeh_2018, Kuntke_Linsner_Steinbrink_Franken_Reuter_2022}. These impacts can be extended for months and have profound social consequences \citep{Andresen_Kurtz_Hondula_Meerow_Gall_2023}. In this context, the operational value of outage prediction models lies in their ability to provide timely, spatially resolved estimates that support repair crew dispatching and reduce the duration and downstream consequences of outage events. Additionally, recent studies have used machine learning-based models to project future outage-related sociodemographic risk under future climate scenarios \citep{guikemaClimateChangeImpacts2025}, while climate change is simultaneously shifting the spatial distribution of extreme events toward regions with little historical exposure \citep{philipRapidAttributionAnalysis2022, kossinPolewardMigrationLocation2014, ciavarellaProlongedSiberianHeat2021, christidisIncreasingLikelihoodTemperatures2020, chemkeClimateChangeShifts2026}. For this reason, it is vital to understand how predictive model accuracy varies across event types and when models are deployed in locations lacking historical outage data. Moreover, the high dimensionality of spatio-temporal weather data poses additional challenges for developing models that generalize reliably in outage location prediction.

In this paper, we evaluate the generalizability of power outage prediction models trained using state-of-the-art AI/ML methods, emphasizing methodological choices in data cleaning and test set construction that can inflate out-of-sample performance estimates. We test the impact of test selection strategies (random, spatial leave-one-one and temporal leave-one-out) and target variables used (relative and absolute) on a the entire U.S. East Coast. To assess the sensitivity of our results to feature selection, we also use the embeddings of the GeoAI foundation model Prithvi WxC \citep{schmude2024prithviwxcfoundationmodel} as input features. Foundation models based on the Visual Transformer architecture (ViT) are at the forefront of state-of-the-art techniques for representation and self-supervised learning of spatial phenomena. These task-agnostic models expand upon Convolutional-based Autoencoders and are trained on very large datasets so they can generalize to a wide variety of use cases. Prithvi WxC is an example of such models, specifically trained on weather and climate data, developed by NASA and IBM, and freely available to use. By design, foundation models aim to learn general-purpose representations, allowing pretrained models to be applied to new tasks with reduced data requirements and potentially improve generalizability.




\section*{Related Work}

\afterpage{
\begin{landscape}
\begin{table}[]
    \centering
    \caption{Recent literature on power outage predictive models and key implementation details. The target variable column shows how the dependent variable is defined. Specifically, we use the acronym $CoP$ (customers out of power) to denote any metric that is an absolute count of customers experiencing outages, while $pCoP$ (percentage of $CoP$) denotes relative (0-1) metrics. Multi-stage methods consist of a classification stage (high-low) followed by a regression model that produces a refined prediction. Lastly, some studies report the performance of a null model for their data, which is labeled as ref.}
    {\renewcommand{\arraystretch}{1.5}
    \begin{tabularx}{1.3\textwidth}{XC{1.2cm}L{2.2cm}XXL{2.5cm}C{2cm}}
\toprule
\textbf{Reference} & \textbf{Static\linebreak Features} & \textbf{Target\linebreak Variable} & \textbf{Test\linebreak Selection} & \textbf{Study Area} & \textbf{Study\linebreak Period} & \textbf{Reported\linebreak Performance} \\
\midrule
\citet{udehProbabilisticStormElectric2024} (\citeyear{udehProbabilisticStormElectric2024})& No & CoP & Manual time range (2017) & 10 NY Counties & 2016 & $0.3$ $R^2$ \\\hline
\citet{aroraProbabilisticMachineLearning2023} (\citeyear{aroraProbabilisticMachineLearning2023}) & Yes & pCoP & Random split & NY, NJ, FL, TX & One event per area & $0.48$ $R^2$ \\\hline
\citet{watsonImprovedQuantitativePrediction2022, cerraiOutagePredictionModels2020, cerraiPredictingStormOutages2019,watsonWeatherrelatedPowerOutage2020, yangEffectLeadTimeWeather2021, yangQuantifyingUncertaintyMachine2020} (\citeyear{cerraiPredictingStormOutages2019}-\citeyear{watsonImprovedQuantitativePrediction2022}) & Yes & log(CoP) & Storm leave-one-out & Manually selected regions in CT, MA and NH & 373 storms & Up to $0.8$ $R^2$ \\\hline
\citet{tervoShortTermPredictionElectricity2019} (\citeyear{tervoShortTermPredictionElectricity2019})& Yes & Damage Level\linebreak (4 classes) & Random split & Two regions in Finland & 2012-2017 & $\approx 70\%$ $F1$ \\\hline
\citet{kabirPredictingThunderstormInducedPower2019} (\citeyear{kabirPredictingThunderstormInducedPower2019}) & Yes & Multi-stage (CoP) & Random split & Specific service regions in AL & 11 storms 2009-2019 & $14.$ $MAE$ ($24.6$ ref) \\\hline
\citet{shashaaniMultiStagePredictionZeroInflated2018} (\citeyear{shashaaniMultiStagePredictionZeroInflated2018})& Yes & Multi-stage (CoP) & 2 manually selected storms & AL, FL & Ivan (2004), Katrina (2005), Dennis (2005) & $\approx 300$ $MAE$ ($800$ ref) \\
\bottomrule
\end{tabularx}
}
    \label{tab:lit-review}
\end{table}
\end{landscape}
\clearpage
}

\subsection*{Power Outage Prediction Models}

Existing methods for power outage prediction can be broadly classified into fragility-based methods and statistical models \citep{fatimaMachineLearningPower2024}. Among the statistical models, several supervised machine learning techniques have been applied to the problem of predicting outages from weather conditions. These include Support Vector Regression \citep{tripathiDownscalingPrecipitationClimate2006}, Tree-based models \citep{nateghiMultiDimensionalInfrastructureResilience2018, cerraiPredictingStormOutages2019,watsonImprovedQuantitativePrediction2022, aroraProbabilisticMachineLearning2023, kabirPredictingThunderstormInducedPower2019} and Artificial Neural Networks \citep{tervoShortTermPredictionElectricity2019, madasthuEnsembleDeepLearning2023, wangDeepLearningBasedWeatherRelated2024, rastgooExtremeOutagePrediction2025}. The most common covariates used include wind speed and wind gust, precipitation, tree coverage, population density, and elevation \citep{shashaaniMultiStagePredictionZeroInflated2018, mcrobertsImprovingHurricanePower2018, aroraProbabilisticMachineLearning2023}. Additional modeling considerations addressing some of the challenges specific to data-driven power outage prediction have been discussed. For instance, power outage datasets are typically imbalanced in favor of data points with zero outages. \citet{shashaaniMultiStagePredictionZeroInflated2018} proposed a multi-stage prediction methodology to address this issue, while \citet{watsonImprovedQuantitativePrediction2022} used SMOTE sampling, which under-samples common examples and synthetically generates new ones from the underrepresented class. Table \ref{tab:lit-review} presents a summary of the most relevant machine learning power outage prediction models and key implementation details for each one.

Across the literature, we observe three modeling choices that we hypothesize impact model generalization: the use of random split test selection, absolute magnitude of outage as the target variable (customers out of power) instead of a relative metric, and heterogeneity in the study area and period selection. First, a random split test set leads to overly optimistic results due to high spatio-temporal correlation in the covariates. Weather, land-use, and other remote sensing data used for power outage prediction are very likely to show spatial dependence structures, meaning nearby observations are going to be more similar than distant observations. If not properly accounted for, these conditions often lead to non-independent error structures and overfitting, which produce inflated performance metrics \citep{robertsCrossvalidationStrategiesData2017, plotonSpatialValidationReveals2020}. Second, the natural link between total population and absolute outage magnitude makes it difficult to generalize trends and interpret error metrics like mean absolute error. Because population levels vary substantially across geographic areas (a county in rural South Carolina may have tens of thousands of residents while one in suburban New Jersey may have hundreds of thousands), the differences between areas in the number of customers affected by an outage tend to be larger than the differences within any single area. This means that a statistical model trained to minimize overall prediction error will prioritize learning which area a data point belongs to, rather than learning the underlying relationship between storm characteristics and outage severity. As a result, the model may perform well on the regions it was trained on but fail to produce reliable predictions when applied to new, unseen areas.

Finally, while many outage events have been used to train and test predictive models, there is limited evidence for how models trained on specific events generalize to other events or areas. \citet{aroraProbabilisticMachineLearning2023} recognized this limitation and trained statistical models (Generalized Linear Models (GLM), Poisson GLM, Random Forest, etc.) with data from over 1,900 cities in the US over three hurricane events. However, they find these models showed signs of overfitting to the local conditions when tested on new regions. In particular, conditions such as wind speeds above those found in training would produce unrealistic results. \citet{watsonWeatherrelatedPowerOutage2020} showed that reducing heterogeneity of weather conditions improved performance and that sufficiently diverse data leads to more generalizable results. 

\subsection*{Use of Foundation Models to improve generalizability}

Recent advancements in Geospatial Artificial Intelligence provide methods to potentially address the generalizability limitations of existing approaches. Foundation models are machine learning models trained on very large datasets using a self-supervised approach \citep{schneiderFoundationModels2024}. The goal of this process is to obtain a model that can produce dense, relatively low-dimensional representations of the input domain. This is often achieved using an encoder-decoder architecture in which an encoder sub-network produces low-dimensional embeddings, and a decoder sub-network reproduces the starting input data from those embeddings. Because foundation models are trained on very large datasets, these representations tend to generalize much better than traditional task-specific models. Foundation models for geospatial applications, in particular, have gained attention in recent years \citep{janowiczGeoFMHowWill2025}. Downstream tasks for these models include predicting geolocations from satellite images \citep{haasPIGEONPredictingImage2024,zhangEarthGPTUniversalMultimodal2024}, enhancing flood inundation mapping \citep{kostejnUPrithviIntegratingFoundation2025}, land surface monitoring \citep{spradlinSatVisionTOAGeospatialFoundation2024}, and spectral imaging segmentation and classification \citep{liS2MAESpatialSpectralPretraining2024, heFoundationModelBasedMultimodal2024}.

Prithvi WxC is a foundation model trained on weather data from the entire globe \citep{schmude2024prithviwxcfoundationmodel}. Transfer learning can be applied to build a power outage predictive model that leverages the dense embedding of weather features generated by the encoder network of Prithvi WxC to produce spatial predictions of power outages. The model was trained using a total of 30 weather features, 4 static features (including land cover), as well as climatological features estimated by computing the deviation from historical climate. All the data is openly available. Because Prithvi WxC was trained on global instances, feature representations from these embeddings are able to capture patterns from heterogeneous regions.

\section*{Methods}

The following section details our methodology. We train the same model, making adjustments to the train and test selection strategies, target variable, and feature set used, in order to evaluate the impact on the generalizability of each of these parameters. In the first part of this section, we describe the weather and outage data sources, followed by the data cleaning procedures applied to reduce zero-inflation in the outage records. We then outline the extraction of spatial embeddings from a GeoAI foundation model, and describe the experimental design used to evaluate the impact on generalizability performance of each methodological decision.

In summary, we train XGBoost models across experiments that vary in three dimensions: input features, output variable, and train-test split strategy. Input features are either ERA5 reanalysis variables (optionally augmented with static tree density) or Prithvi WxC embeddings, aggregated at the county-day level. The output variable is either absolute customer outage counts or the percentage of customers affected. For the train-test split, we consider random 80-20, spatial leave-one-state-out, and temporal leave-one-event-out strategies. All experiments were conducted using XGBoost as it is well-suited for the scale of data available, and tree-based models are the dominant architecture in the literature of power outage prediction \cite{nateghiComparisonValidationStatistical2011, cerraiPredictingStormOutages2019}. While neural network architectures would enable more sophisticated approaches, such as fine-tuning Prithvi WxC or capturing spatial correlations via Graph Neural Networks, they demand substantially greater data and computational resources and are thus left as future research as data availability improves.

\subsection*{Data Sources}
\subsubsection*{Weather Reanalysis and Tree Density features}
The European Center for Medium-Range Weather Forecast (ECMWF) provides a series of geo-referenced weather variables datasets. In particular, we used the ERA5 reanalysis dataset as the baseline source of weather information for the study area. The ERA5 reanalysis dataset provides hourly information on a wide range of atmospheric features at a resolution of 31 km (roughly 0.25 latitude/longitude degrees) from 1940 to the present. \citep{C3S_2018}. We selected a set of 3 features from the ERA5 single-level dataset and 7 features from the ERA5 land dataset. Table \ref{tab:regular_features} enumerates all the features used.

Tree density is commonly used as a static covariate in power outage predictive models \citep{guikemaPredictingHurricanePower2014, guikemaModelingPowerOutage2018, shashaaniMultiStagePredictionZeroInflated2018} because it is hypothesized that many outages are caused by trees falling on distribution infrastructure. We collected tree density information as a rasterized layer from the NLCD database. Additionally, because tree density is only relevant for areas with distribution infrastructure, we used the information in the other layers of the NLCD to determine which pixels of the raster of tree density were relevant. For instance, we kept all the pixels of tree density where the NLCD reported urban development and discarded the rest. We then aggregated these values by county using the polygon representation of each area and averaged the selected pixels.

\begin{table}[t]
    \centering
    \caption{Features used to train the predictive models. All features were aggregated to a county-day resolution using an average spatial join operation between point features and county polygons. The max function was used for temporal aggregation.}
    \begin{tabular}{l|c|c}
         \textbf{Feature} & \textbf{Abbreviation} & \textbf{Source} \\\midrule
         Temperature Dew Point & \texttt{d2m} & ERA5 - Single Levels\\
         Temperature & \texttt{t2m} & ERA5 - Single Levels\\
         Wind Gust & \texttt{i10fg} & ERA5 - Single Levels\\
         Mean Large-Scale Precipitation Rate & \texttt{mlspr} & ERA5 - Land\\
         Wind Speed & \texttt{u10} and \texttt{v10} & ERA5 - Land\\
         Volumetric Soil Water & \texttt{swvl1-4} & ERA5 - Land\\
         Tree Density $\times$ Urban Development & \texttt{tree\_dens} & NLCD\\\bottomrule
    \end{tabular}
\label{tab:regular_features}
\end{table}

\subsubsection*{Power Outage Database}

Power outage data was acquired from the PowerOutage.us \citep{poweroutageus} database. This dataset consists of county-level power outage records as reported by utility websites. Data records are available with inconsistent temporal resolution, and the values are disaggregated by utility. The sum over the maximum value among all utilities for each county on each day was used as the true value of outages in a county-day pair. Additionally, the column labeled "Customers Tracked" was used as a reference for the total population served by the utility. This value is used in the experiments where the percentage of the population without power is the target feature, instead of the absolute value of customers without power.

\subsection*{Data Cleaning}

The main goal of the data cleaning process was to reduce the zero-inflation in the data due to most counties not experiencing significant outages most of the days. There are several methods to achieve this in the literature. The prevailing one is to manually select dates for which it is known in advance that there were significant weather-related outages \citep{aroraProbabilisticMachineLearning2023, kabirPredictingThunderstormInducedPower2019, shashaaniMultiStagePredictionZeroInflated2018}. One noteworthy exception is the use of basic signal matching procedures to detect dates when both weather and the number of outages were outliers in \citep{udehProbabilisticStormElectric2024}. We used a simple outlier detection procedure for the number of outages on each day. This decision was motivated by the goal of minimizing arbitrary decisions and preserving the applicability of the procedure in general circumstances. In particular, for each county, we flagged a date as an outlier if the peak number of outages on that date was higher than the moving average by three standard deviations. The window size for the moving average was chosen to be one year. Then, we selected all the dates for which there were at least 40 counties with an outlier number of outages. The value of 40 was chosen based on the values for named events like Hurricane Florence (56 outliers) and Hurricane Michael (175 outliers). This value needs to be adjusted to match the magnitude of the number of counties in the study area if a different dataset is used. We note that this method also identifies significant outage events that may not be named, and therefore, it is less subject to biases. After cleaning the data using this procedure, 28 dates were selected, and $27\%$ of the remaining county-date data points corresponded to at least $5\%$ of the population of the county without power. The resulting data points represent the universe from which each experiment draws train-validation-test splits.

\subsection*{Extracting GeoAI embeddings}
\begin{figure}
    \centering
    \includegraphics[width=.95\linewidth]{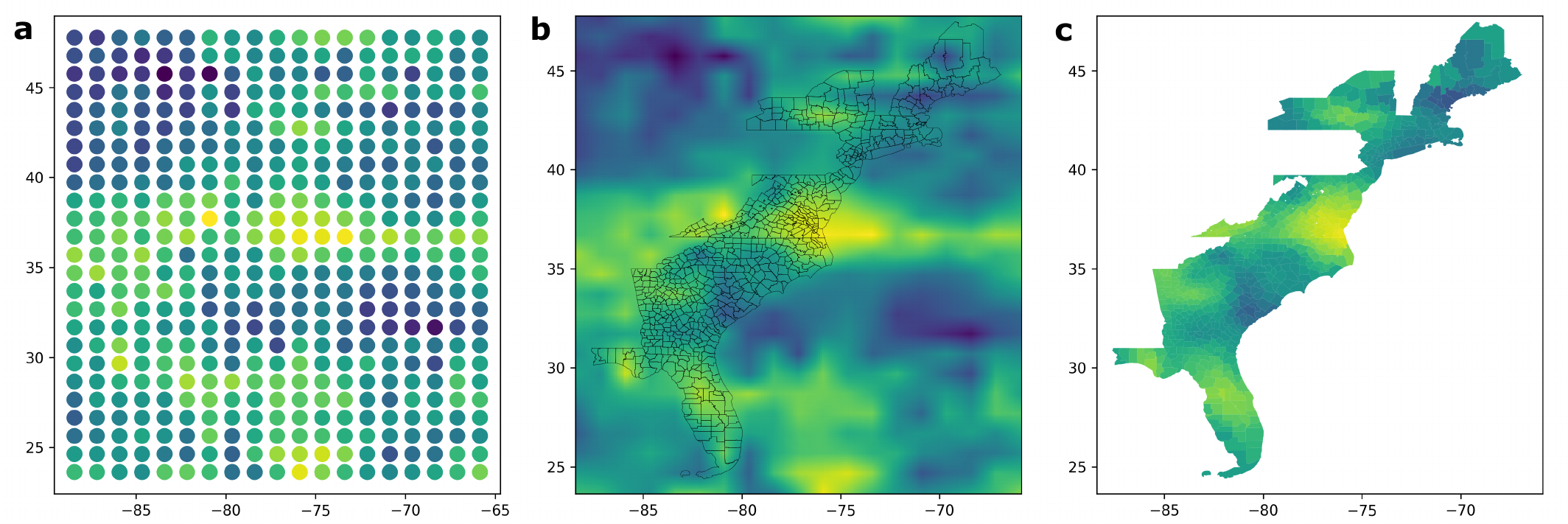}
    \caption{Example of interpolation procedure. (a) Prithvi WxC embedding dimension 4 at original resolution, (b) bilinear interpolation of embedding that covers the extent of the study area, and (c) county-day average for this sample computed as the mean of the interpolated values.}
    \label{fig:interpolation}
\end{figure}

Foundation models produce transferable feature representations that can be adapted to many downstream tasks, making them valuable for improving generalizability in tasks of a common domain. Many foundation models, including Prithvi WxC, follow an encoder-decoder architecture. The main feature of this kind of model is the embedding representation between the encoder and the decoder network, which contains an efficient representation of the input data and can be used as the input for a downstream model. In order to leverage these embeddings, we downloaded MERRA-2 reanalysis data \cite{gelaroModernEraRetrospectiveAnalysis2017} for the period of interest following the structure provided by the developers of Prithvi WxC. Then, we ran the encoder of the model on that data and extracted, for each timestamp, the embeddings corresponding to the bounding box of the states on the East Coast of the US. The last step was to map the regular grid of embeddings to the polygon representing each county, which was performed by applying bilinear interpolation and then a raster mask. Figure \ref{fig:interpolation} shows an example of one embedding dimension of Prithvi WxC for one timestamp, as well as the result of applying bilinear interpolation and the raster mask to obtain county-timestamp values. We tested the impact of using the embeddings of Prithvi WxC as a replacement for the manually selected ERA5 and NLCD features. The advantage of this approach is that the embeddings of the foundation model contain significantly more information and can potentially improve model generalizability because it was trained on global data from a large time period.

Prithvi WxC was trained on the MERRA-2 weather reanalysis dataset for the entire planet from 1980 to 2023. Both the trained model and the MERRA-2 dataset are openly accessible.  The embeddings of Prithvi WxC are 2560-dimensional vectors for each cell in a regular grid with about 50-60 km of spatial resolution and 3-hour temporal resolution, meaning for each county-timestamp pair, we have a resulting vector of 2560 features. Additionally, the model expects the MERRA-2 features for the current timestamp, as well as a previous one and a prediction lead time. We used 12:00 AM and 12:00 PM as input times for each relevant date and set the lead time to zero hours to represent embeddings of the given input features and not of any future time. More details about the significance of these parameters are found in the original Prithvi WxC publication \citep{schmude2024prithviwxcfoundationmodel}.

\subsection*{Experiment design}

\begin{figure}[]
    \centering
    \includegraphics[width=.95\linewidth]{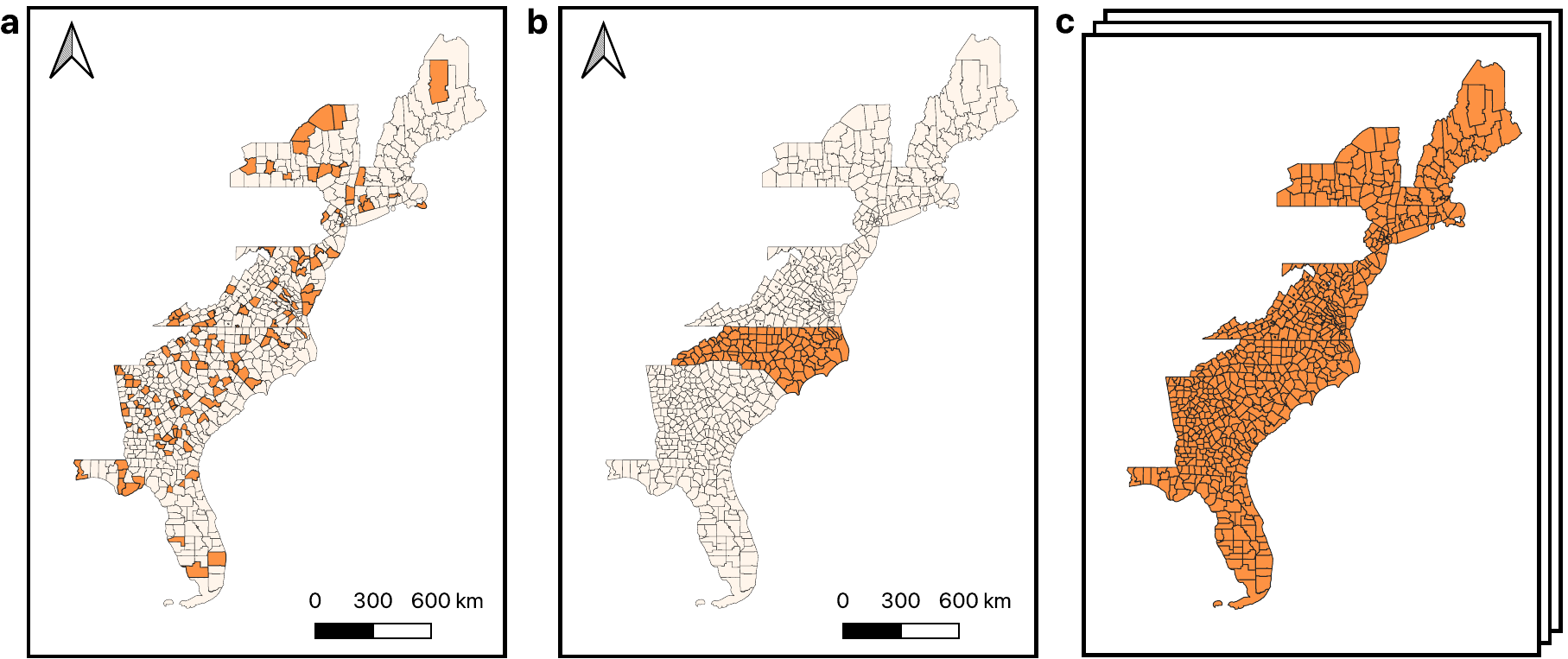}
    \caption{Strategies for selecting a test set. In each subfigure, an example for each strategy is provided. Counties in orange would be chosen as the test set. (a) Example of random split strategy using 80-20 ratio, (b) Leave-state-out strategy, North Carolina in this case, and (c) Leave-event-out, which selects the entire study region as test set, for a specific date.}
    \label{fig:sample_plot}
\end{figure}

As mentioned above, the objective of these experiments is to study the generalizability impacts of (1) using a random sample with spatially correlated data, (2) using absolute metrics of outage magnitude, and (3) using data for a wide range of event types and locations. We present our results in three experiments differentiated by test set selection strategy, namely, random 80-20 split, spatial leave-one-out (LOO), and temporal LOO. For spatial LOO, we used each of the 13 states included in the data as a test set, and for temporal LOO, we used each event date as a different test set. Figure \ref{fig:sample_plot} shows one example of each strategy. This allows us to answer question (1). Inside each experiment, we present performance results for models trained to predict total population out of power ($CoP$) and percentage of population out of power ($pCoP$) in order to answer question (2). Lastly, in the spatial LOO and temporal LOO experiments, we disaggregated results based on which state or event was left as the test set to explore generalizability patterns and answer question (3). As mentioned in the Data Cleaning section, we used an intentionally permissive procedure to clean the data that would allow us to evaluate predictive performance for as many events of different types as possible. In each experiment, we present results for models trained with reanalysis and Prithvi WxC embeddings to evaluate the performance of different feature sets.

We used a validation set formed from a random sample from the training set in each experiment to perform early stopping and avoid overfitting. Based on preliminary results, we maintained most of the parameters at their default values, with the exception of maximum tree depth, which was set to 8 instead of 6, and a lambda and alpha values of 2 and 0.5, respectively. In the case of the random split strategy, we performed stratified random sampling to maintain the proportion of significant outages in both training and testing sets. The remaining data not used for testing was used to train a model using 5-fold cross-validation, and the 20\% holdout for validation set in each fold was used to determine model size (number of estimators in the case of XGBoost). In a general case, this configuration can be used to tune any number of relevant parameters of the model.


\section*{Results}

In this Section, we present the results of our experiments for each configuration. The acronyms $CoP$ and $pCoP$ were used to refer to experiments trained to learn the total customers out of power (absolute metric) and the percentage of customers out of power (relative metric), respectively.

\subsection*{Best-case scenario performance: Random split test set}

We first evaluated model performance under a random 80-20 train-test split, treating this as a best-case scenario baseline and as a representation of the bulk of the literature. Because this approach ignores the spatio-temporal correlations inherent in weather covariates, training and test sets will inevitably contain very similar samples, which systematically inflate performance metrics and make it unrealistic for operational forecasting. Each configuration was run 20 times with different test samples to account for incidental sampling biases.

\begin{figure}[]
    \centering
    \includegraphics[width=.75\linewidth]{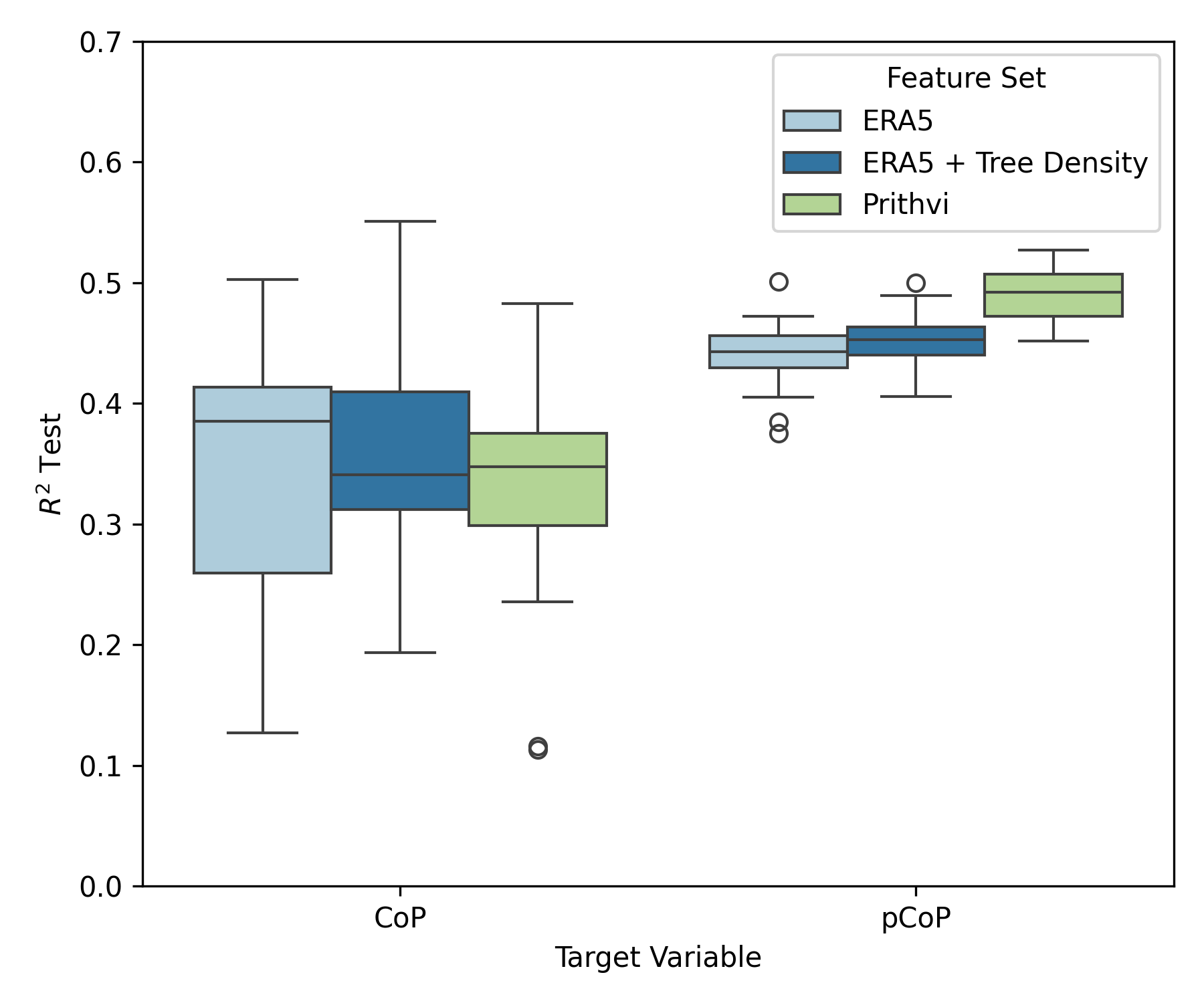}
    \caption{Performance of random split experiment for each feature set and target variable as percentage of variance in the dependent variable that is accounted for in the model ($R^2$).}
    \label{fig:split_boxplot}
\end{figure}

Figure \ref{fig:split_boxplot} shows the $R^2$ of each configuration across different test set samples, and Table \ref{tab:split_results} presents the 95\% confidence intervals for each distribution of $R^2$ and Mean Absolute Error. Predictive accuracy under the random 80-20 split yields approximately 45\% $R^2$ for the relative outage metric (pCoP) and about 33\% for the absolute metric (CoP). These values fall slightly below those reported in the literature (\citet{aroraProbabilisticMachineLearning2023} use the same dataset but a smaller region), which may be explained by the larger spatial domain of our experiments (entire east coast compared to a couple of states), reduced influence of incidentally beneficial samples due to repeated experiments, and differences in feature set composition.

We find no substantial performance variations across different feature sets. Prithvi embeddings provide better predictions for relative outage magnitude, though at the cost of higher dimensionality and data processing complexity. Conversely, ERA5 features combined with static covariates perform moderately better for absolute outage magnitude. We believe the introduction of static covariates makes it possible to identify a county, and therefore, this increase in performance can be explained by the capacity of the data to represent inter-county variability. This effect is more pronounced when predicting the absolute number of outage counts because the target is naturally correlated with county population, whereas using the  percentage of customers without power mitigates this issue.

Lastly, model performance for absolute outage magnitude is notably sensitive to the specific random test set. In some splits, $R^2$ falls below 20\%, which is likely a product of whether the sampled test counties include those with high variability in outage count. 

\begin{table}[]
    \centering
    \caption{95-percent confidence interval for $R^2$ and mean absolute error ($MAE$) performance metrics for each experiment in the 80-20 split category.}
    \begin{tabular}{lcc|cc}
    \toprule
    & \multicolumn{2}{c}{$R^2$} & \multicolumn{2}{c}{$MAE$} \\
    & pCoP & CoP & pCoP & CoP\\
    \midrule
    ERA5 & 0.44 ± 0.01 & 0.32 ± 0.07 & 0.06 ± 0.00 & 1,807.20 ± 48.13  \\
    ERA5 + Tree Density & 0.45 ± 0.01 & \textbf{0.36} ± 0.04 & 0.06 ± 0.00 & 1,796.08 ± 47.62 \\
    Prithvi & \textbf{0.49} ± 0.01 & 0.33 ± 0.04 & \textbf{0.05} ± 0.00 & \textbf{1,720.44} ± 32.31 \\
    \bottomrule
    \end{tabular}
    \label{tab:split_results}
\end{table}

\subsection*{Spatial generalizability performance}

\begin{figure}[t]
    \centering
    \includegraphics[width=.95\linewidth]{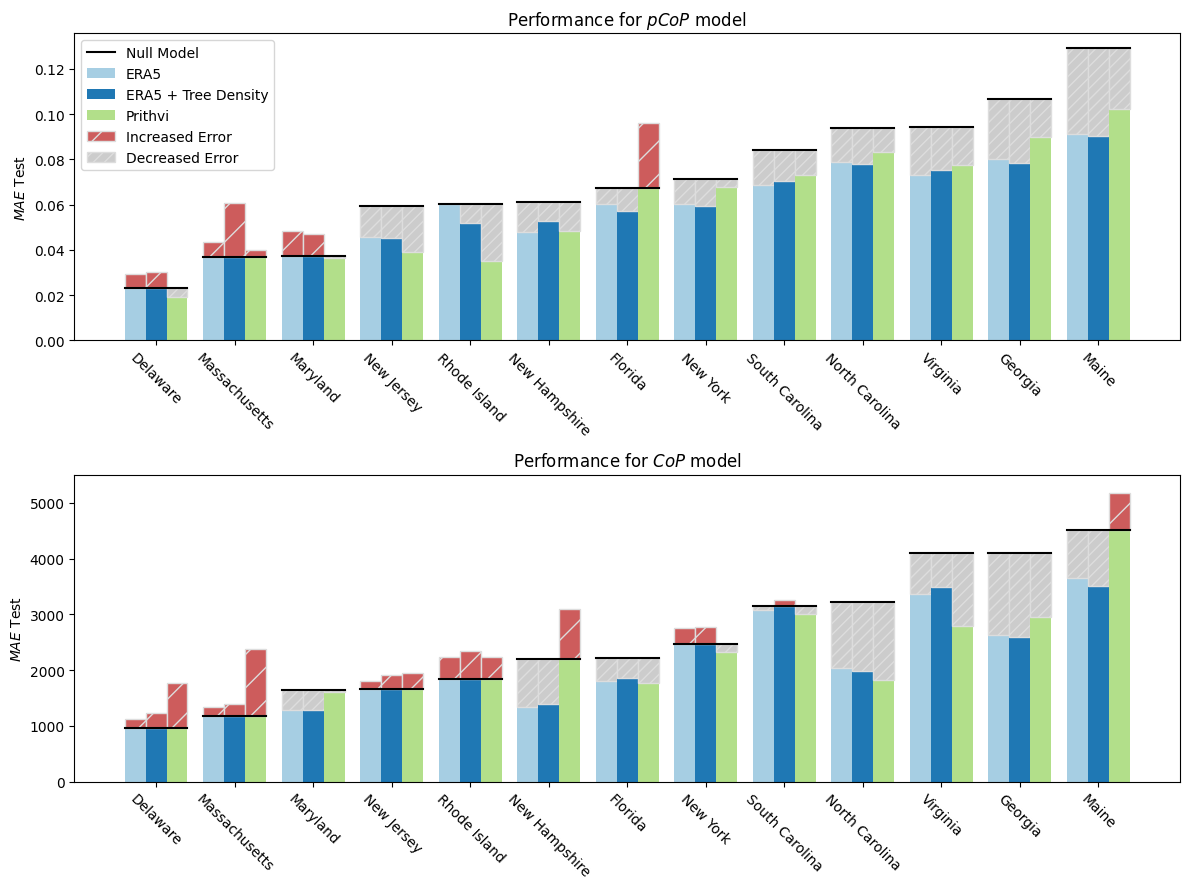}
    \caption{Mean absolute error for the leave-state-out experiment for each target variable ($pCoP$ and $CoP$), feature set and state. Lower is better. In each case, $MAE$ of using a null model is labeled, and the positive or negative difference between the weather model and the null model is presented in grey or red, respectively.}
    \label{fig:spatial_mae}
\end{figure}

In order to assess the generalizability of a power outage prediction model across regions, we train the model following the same training and validation procedure but completely leaving out one state at a time, which is then used as a test set. Figure \ref{fig:spatial_mae} presents the results for each model in terms of mean absolute error and in context with the performance of a null model, which naively predicts the average of the test set samples each time.

The average $MAE$ (across all states and feature sets) is $0.061$ and $2,317$ for $pCoP$ and $CoP$ respectively, compared to the best-case scenario model of around $0.06$ and $1,800$ of the best-case scenario model (Table \ref{tab:split_results}). The larger difference in error for the absolute metric can be explained by the importance of population size for the absolute magnitude of outage prediction. When the training data does not contain information about historical patterns for that particular area, it is difficult to estimate the absolute magnitude of outages due to this inherent correlation.

While predictions of relative outage magnitude are close on average to the best-case scenario, we find performance varies across states. Importantly, in many cases, the trained models do worse (higher error) than the null model, meaning the patterns learned by the model do not correspond with the patterns of correlation between covariates and the target variable represented in the test set. We find that this is consistent for states like Delaware and Massachusetts, and is more sensitive to other training conditions for Maryland, New Jersey, and Rhode Island. These are also areas where prolonged outage events are not historically linked to hurricanes, which could explain the poor performance. We explore the event type dimension in the next section.


In terms of feature selection, models trained on Prithvi embeddings are slightly more consistent in terms of generalizable patterns. For $pCoP$ models, Florida is the only state where Prithvi embeddings capture incorrect patterns. However, performance is still poor across all experiments, which is evidence that the selection of the feature set is not a limiting factor in model generalizability.


\subsection*{Temporal generalizability performance}

We evaluate the temporal generalizability of machine learning models for power outage prediction by excluding entire outage events from the training set. This leave-one-event-out design directly assesses model performance on unseen events. Each event is additionally categorized by extreme weather type as Hurricane, Winter Storm, or Other, where the latter primarily includes unnamed tornadoes and thunderstorm events. Because events are identified through an unsupervised procedure, some events classified as Other may reflect widespread outages not directly attributable to weather. However, this is unlikely as we require at least 40 counties to experience abnormally high outage levels on the same day for an event to be considered significant (see Methods).

\begin{figure}[]
    \centering
    \includegraphics[width=.95\linewidth]{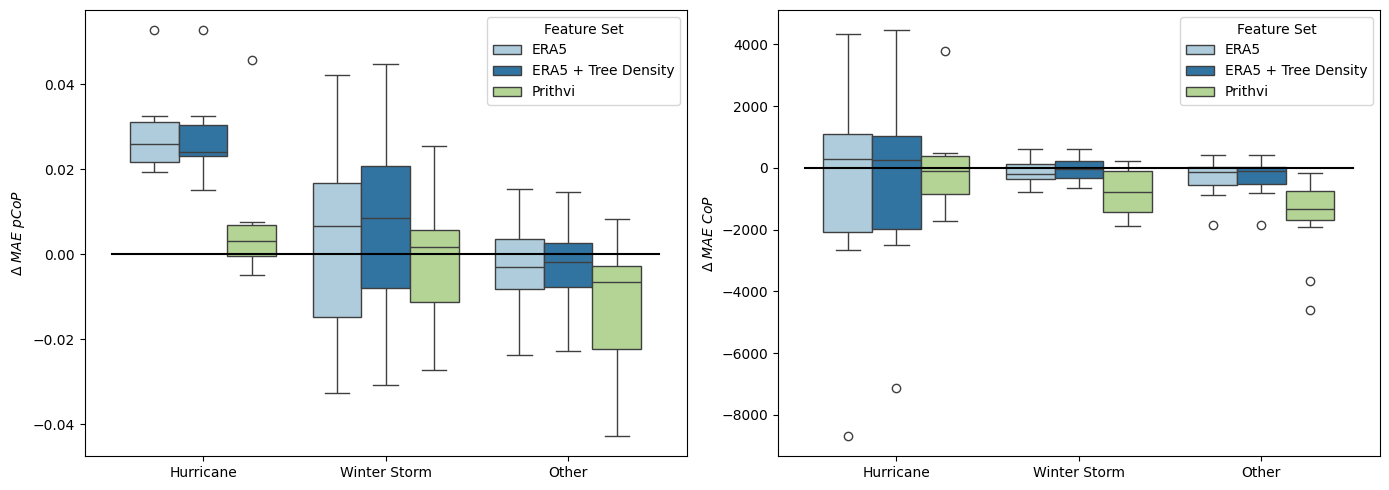}
    \caption{Results of the leave-event-out experiment as the difference in Mean Absolute Error between the trained model and null model. Higher is better.}
    \label{fig:temporal_features}
\end{figure}

This experimental setup closely resembles real-world deployment conditions, where no prior information about the infrastructure impacts of the ongoing event is available at prediction time. Figure \ref{fig:temporal_features} summarizes the performance improvement over a null model for each feature set, grouped by event type, using both $pCoP$ and $CoP$ as target variables. Across all event types, none of the evaluated models meaningfully outperform the null model when predicting the absolute number of customers out of power ($CoP$). This result highlights the high variability in outage magnitude across events and the difficulty of predicting absolute impacts. Moreover, because different events predominantly affect different regions, the underlying population at risk varies substantially, further complicating absolute magnitude prediction. For example, outages concentrated in urban areas of Florida versus those affecting large portions of Virginia would show very different total outage numbers.

\begin{figure}[]
    \centering
    \includegraphics[width=.95\linewidth]{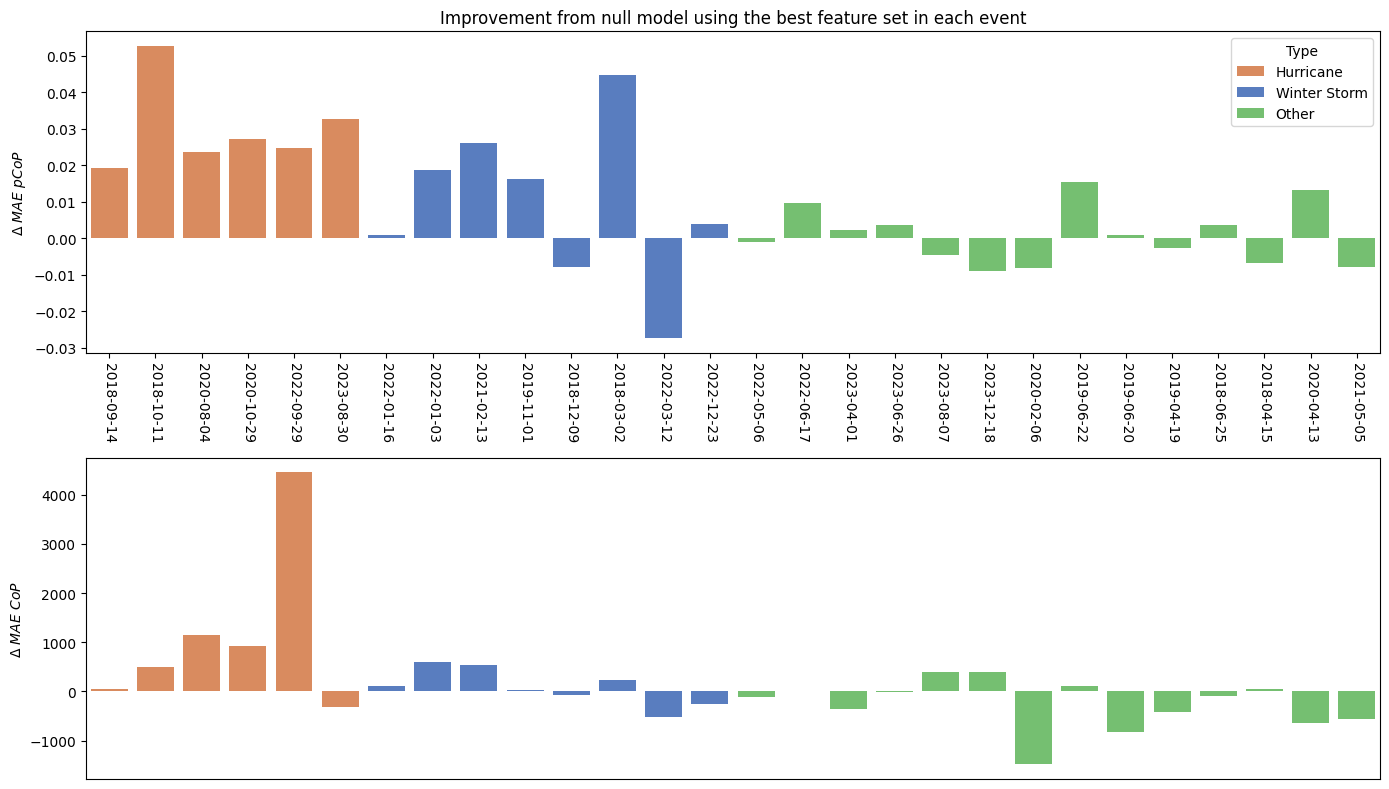}
    \caption{Results of the leave-event-out experiment as difference in Mean Absolute Error between null model and the best trained model among all feature sets in each case, colored by event type. Higher is better.}
    \label{fig:temporal_events}
\end{figure}

In the case of relative outage magnitude ($pCoP$) prediction, we observe that Hurricane events are the only category for which impacts can be consistently predicted with improved accuracy over the null model. Within this subset, models using Prithvi embeddings perform significantly worse than those based on ERA5 features. This finding suggests that, while the foundation model embeddings may support spatial generalization, the patterns they encode do not transfer effectively to unseen events. One plausible explanation is the high dimensionality of the Prithvi embeddings, which may induce overfitting during training.

Figure \ref{fig:temporal_events} presents the performance of the best-performing model for each event, colored by event type. The feature set associated with each best-performing model and the corresponding error metrics are reported in Appendix Table \ref{tab:appendix_temporal}. These results further confirm that outage impacts from Hurricane events are more predictable and that the weather patterns driving these outages exhibit greater temporal generalizability. In contrast, Winter Storms and Other event types do not display comparable predictability. Although most Hurricane events achieve reasonable predictive performance, the optimal feature set varies substantially across events. This variability explains why, in Figure \ref{fig:temporal_features}, no single feature set consistently improves upon the null model in the $CoP$ case, even within the Hurricane category.

\subsection*{Summary of experimental design}

Table \ref{tab:summary} summarizes our experimental design and presents the Mean Absolute Error (MAE) for each experiment type. We recommend that future research on power outage prediction models report similar summaries to address potential issues due to spatio-temporal correlation in covariates, thereby improving transparency and reproducibility.

We include in this table the dimensions we have found that impact the generalizability of power outage prediction models. First, the use of an absolute or relative target variable, which is an important consideration due to the conflation of outage magnitude and total population. Previous work has used logarithms of these quantities, in which case, we believe it is also important to disclose. Then, we describe the feature sets tested, and highlight in particular the source and whether or not static features are included. We do not believe testing different feature sets is a requirement for future studies, but testing the impact of static features should be part of the experimental design. Lastly, we include different testing strategies, specifically leaving out samples based on the spatial and temporal domains. This is critical to assess the generalizability of the results presented, and the number of independent test sets formed from this selection is also a careful consideration when evaluating the results.

\begin{table}[!h]
    \centering
    \caption{Summary of experimental design and average Mean Absolute Error. Different test types produce different number of test sets. Spatial Leave-One-Out experiments are conducted by leaving all the data from a single State out of training and used for testing. Temporal Leave-One-Out experiments leave data for a given date out of training and used for testing.}
    \label{tab:summary}
    \begin{tabular}{lll|cc}
\toprule
\textbf{Target Variable} & \textbf{Feature Set} & \textbf{Test Type} & \textbf{\# of test sets} & \textbf{Avg. Test MAE} \\
\midrule
\multirow[c]{9}{*}{CoP (absolute)} & \multirow[c]{3}{*}{ERA5 \textit{(No static features)}} & 80-20 Split & 20 & 1,807.202 \\
 &  & Spatial LOO & 13 & 2,185.835 \\
 &  & Temporal LOO & 28 & 2,766.400 \\
\cline{2-5}
 & \multirow[c]{3}{*}{ERA5 + Tree Density \textit{(Static features)}} & 80-20 Split & 20 & 1,796.083 \\
 &  & Spatial LOO & 13 & 2,235.405 \\
 &  & Temporal LOO & 28 & 2,674.593 \\
\cline{2-5}
 & \multirow[c]{3}{*}{Prithvi \textit{(Embedded static features)}} & 80-20 Split & 20 & 1,720.445 \\
 &  & Spatial LOO & 13 & 2,530.455 \\
 &  & Temporal LOO & 28 & 3,347.115 \\
\midrule
\multirow[c]{9}{*}{pCoP (relative)} & \multirow[c]{3}{*}{ERA5 \textit{(No static features)}} & 80-20 Split & 20 & 0.060 \\
 &  & Spatial LOO & 13 & 0.061 \\
 &  & Temporal LOO & 28 & 0.078 \\
\cline{2-5}
 & \multirow[c]{3}{*}{ERA5 + Tree Density \textit{(Static features)}} & 80-20 Split & 20 & 0.059 \\
 &  & Spatial LOO & 13 & 0.061 \\
 &  & Temporal LOO & 28 & 0.076 \\
\cline{2-5}
 & \multirow[c]{3}{*}{Prithvi \textit{(Embedded static features)}} & 80-20 Split & 20 & 0.055 \\
 &  & Spatial LOO & 13 & 0.062 \\
 &  & Temporal LOO & 28 & 0.088 \\
\bottomrule
\end{tabular}
\end{table}

\section*{Discussion and conclusion}

This study evaluated the spatiotemporal generalizability of machine-learning-based power outage prediction models under data availability conditions that compare to realistic deployment scenarios. Across a range of feature sets, target definitions, and test selection strategies, our results show that predictive performance degrades substantially when models are evaluated on unseen regions or unseen events. While the widely used random train-test splits yield model accuracy comparable to that reported in recent studies, these results are largely inflated by spatial and temporal autocorrelation in the covariates and do not reflect real-world forecasting conditions. Under leave-one-state-out and leave-one-event-out experiments, performance improvements over a simple null model are typically small, highly variable, and in many cases not statistically meaningful, limiting the operational value of these models for disaster preparedness and response. Accordingly, we conclude that the generalizability of power outage prediction models using random train-test splits is poor due to data leakage from training to test sets.

The limited generalizability observed in our experiments has important implications for operational planning. Even when average errors appear acceptable, the large variability across states and events implies that model outputs would be unreliable. Temporal generalization shows especially poor performance. When entire events are excluded from training, none of the evaluated models consistently outperform a null baseline for predicting absolute outage magnitude, and only for hurricane events, the models show modest improvements when predicting relative impacts. This sensitivity to event type suggests that outage-weather relationships learned from historical data do not readily transfer across heterogeneous conditions, undermining confidence in model-based decision support during future events.

GeoAI foundation model embeddings offer a limited improvement. The use of Prithvi WxC embeddings produced more consistent performance than hand-picked features in some spatial generalization settings, especially for relative outage metrics. However, these gains are modest and do not extend to temporal generalization, which is arguably the more relevant challenge for forecasting. Moreover, leveraging GeoAI embeddings has practical costs associated with extracting and storing high-dimensional representations of weather conditions. Significant computational and memory resources are needed, while the resulting features are difficult to interpret. These trade-offs weaken the case for deploying such models in operational environments, particularly when performance gains are marginal.

Data limitations also represent a meaningful constraint in expected accuracy. Weather reanalysis products, while attractive due to their global coverage and consistency, are not designed to optimally represent extreme conditions that drive infrastructure failures. Because they combine real observations with physical models, localized extremes that are critical for outage prediction may be smoothed out. Although some prior studies have achieved stronger results using specialized wind-field models \cite{guikemaPredictingHurricanePower2014, shashaaniMultiStagePredictionZeroInflated2018}, these approaches often rely on event-specific calibration, limited spatial domains, or data sources that are not readily available for real-time forecasting. As a result, their scalability and practical applicability are highly dependent on the quality and quantity of data available in each case.

We believe the observed limitations are unlikely to be resolved by substituting alternative machine-learning architectures for XGBoost. The high performance under best-case scenario experiments, and training and validation set $R^2$ consistently above $90\%$ and $30\%$ in all of our experiments, suggests that models are already capable of fitting the available data. Given the relatively small number of independent extreme events and the strong spatiotemporal dependence in the data, more complex architectures would likely exacerbate this issue rather than improve generalization. However, if data availability improves, neural network models could be a better fit for this problem than tree-based architectures because they make fine-tuning Prithvi WxC possible and because spatial correlations and network-like structures (like the distribution grid) can be directly accounted for in the model architecture. High-dimensional embeddings further disadvantage neural networks in this context because data requirements grow exponentially with dimensionality for neural network models, and small datasets increase the risk of overfitting and vanishing gradients. These challenges are less prominent when using tree-based models because of their implicit feature selection at each split.

Future research could also explore multi-stage prediction, where a first-pass model routes each sample to a specialized sub-model trained for specific conditions. This architecture would relieve any single model from handling the full range of weather variability, potentially improving precision on rare or extreme events. The key challenge in that approach lies in guaranteeing data availability for all stages and sensitivity to the partitioning criteria.

Our findings are consistent with earlier work showing that high predictive performance is achievable primarily when models are trained on large, region-specific, and often proprietary datasets. Utilities with access to detailed infrastructure inventories, high-resolution scale outage logs, and localized damage reports may be able to develop accurate models for their own service territories. However, such conditions do not generalize to publicly available data or to regions with sparse historical exposure to extreme events. Compounding this issue, utilities have limited incentives to publicly report detailed outage information, leading to incomplete and heterogeneous datasets that further impede model transferability.

Overall, this work demonstrates that current data-driven outage prediction models, when trained on publicly available data and evaluated under realistic deployment scenarios, offer limited and uncertain benefits for operational decision-making. Progress in this area will likely require a shift away from marginal architectural improvements toward better data coverage, more open reporting standards, and evaluation protocols that explicitly reflect the conditions under which models are intended to be used.


\section*{Appendix}

Table \ref{tab:appendix_temporal} presents a comprehensive list of the features that produced the best performance for each event in our dataset, as well as the corresponding mean absolute error for both relative and absolute outage magnitude metrics.

\begin{table}[!ht]
    \centering
    \caption{Detailed results of the leave-event-out experiment for the best feature set in each run.}
    \label{tab:appendix_temporal}
    \begin{tabular}{lll|lc|lc}
        \toprule
        \multirow{2}{*}{Type} & \multirow{2}{*}{Date} & \multirow{2}{*}{Name} & \multicolumn{2}{c}{CoP} & \multicolumn{2}{c}{pCoP} \\
         &  &  & Best Features & MAE & Best Features & MAE \\
        \midrule
        Hurricane & 2018-09-14 & Florence & Prithvi & 3,060.08 & ERA5 & 0.09 \\
        Hurricane & 2018-10-11 & Michael & Prithvi & 4,028.98 & ERA5 + Tree Density & 0.17 \\
        Hurricane & 2020-08-04 & Isaias & ERA5 & 3,452.35 & ERA5 + Tree Density & 0.11 \\
        Hurricane & 2020-10-29 & Zeta & ERA5 & 4,388.31 & ERA5 & 0.08 \\
        Hurricane & 2022-09-29 & Ian & ERA5 + Tree Density & 6,448.92 & ERA5 & 0.05 \\
        Hurricane & 2023-08-30 & Idalia & ERA5 & 2,557.34 & ERA5 + Tree Density & 0.09 \\\midrule
        Winter Storm & 2018-03-02 & Riley & Prithvi & 2,050.21 & ERA5 + Tree Density & 0.11 \\
        Winter Storm & 2018-12-09 & Diego & ERA5 + Tree Density & 2,073.66 & ERA5 + Tree Density & 0.07 \\
        Winter Storm & 2019-11-01 & Unnamed & ERA5 + Tree Density & 1,674.42 & ERA5 + Tree Density & 0.06 \\
        Winter Storm & 2021-02-13 & Uri & ERA5 + Tree Density & 1,419.30 & ERA5 + Tree Density & 0.07 \\
        Winter Storm & 2022-01-03 & Frida & ERA5 & 2,386.17 & ERA5 + Tree Density & 0.09 \\
        Winter Storm & 2022-01-16 & Izzy & ERA5 + Tree Density & 1,444.60 & ERA5 + Tree Density & 0.06 \\
        Winter Storm & 2022-03-12 & Unnamed & ERA5 + Tree Density & 1,734.83 & Prithvi & 0.08 \\
        Winter Storm & 2022-12-23 & Elliott & ERA5 + Tree Density & 2,508.34 & Prithvi & 0.08 \\\midrule
        Other & 2018-04-15 & Unnamed & ERA5 + Tree Density & 1,507.71 & ERA5 + Tree Density & 0.06 \\
        Other & 2018-06-25 & Unnamed & ERA5 & 797.63 & ERA5 & 0.05 \\
        Other & 2019-04-19 & Unnamed & ERA5 & 1,796.39 & ERA5 + Tree Density & 0.07 \\
        Other & 2019-06-20 & Unnamed & ERA5 + Tree Density & 1,999.54 & Prithvi & 0.06 \\
        Other & 2019-06-22 & Unnamed & ERA5 & 963.86 & ERA5 & 0.05 \\
        Other & 2020-02-06 & Unnamed & Prithvi & 2,581.57 & Prithvi & 0.07 \\
        Other & 2020-04-13 & Unnamed & ERA5 + Tree Density & 3,519.10 & ERA5 + Tree Density & 0.09 \\
        Other & 2021-05-05 & Unnamed & ERA5 + Tree Density & 1,296.01 & ERA5 + Tree Density & 0.06 \\
        Other & 2022-05-06 & Unnamed & ERA5 + Tree Density & 821.87 & ERA5 + Tree Density & 0.04 \\
        Other & 2022-06-17 & Unnamed & ERA5 + Tree Density & 1,750.36 & ERA5 & 0.07 \\
        Other & 2023-04-01 & Unnamed & ERA5 & 1,485.34 & ERA5 & 0.05 \\
        Other & 2023-06-26 & Unnamed & ERA5 + Tree Density & 1,340.19 & ERA5 & 0.05 \\
        Other & 2023-08-07 & Unnamed & ERA5 & 2,022.65 & ERA5 & 0.06 \\
        Other & 2023-12-18 & Unnamed & ERA5 + Tree Density & 2,079.94 & Prithvi & 0.07 \\
        \bottomrule
    \end{tabular}
\end{table}

\section*{Acknowledgments}

This work was supported by the DOE American Made Digitizing Utilities Prize.

\section*{Author contributions statement}

Y.E and B.R. formulated the research goals and scope. Y.E and R.V. conceived the experiments. Y.E conducted the experiments.  Y.E. analyzed the results and wrote the manuscript. All authors reviewed the manuscript. 




\bibliographystyle{unsrtnat}
\bibliography{main}

\end{document}